\documentclass{article}

\usepackage{PRIMEarxiv}

\usepackage[utf8]{inputenc} 
\usepackage[T1]{fontenc}    
\usepackage{hyperref}       
\usepackage{url}            
\usepackage{booktabs}       
\usepackage{amsfonts}       
\usepackage{nicefrac}       
\usepackage{microtype}      
\usepackage{lipsum}
\usepackage{fancyhdr}       
\usepackage{graphicx}       
\graphicspath{{media/}}     
\usepackage{textcomp}
\usepackage{amsmath,amssymb,amsfonts}
\usepackage{lipsum}

\usepackage{algorithmic}
\usepackage{xcolor}
\usepackage{bm}
\usepackage{booktabs}
\usepackage { multirow }
\usepackage{adjustbox}

\title{Enhancing Physical Human-Robot Interaction: Recognizing Digits via Intrinsic Robot Tactile Sensing
\thanks{\textit{\underline{Citation}}: 
\textbf{Authors. Title. Pages.... DOI:000000/11111.}} 
}

\author{
  Teresa Sinico, Giovanni Boschetti \\
  Dep. Management and Engineering \\
  University of Padua \\
  Vicenza, Italy\\
  \texttt{\{teresa.sinico\}@phd.unipd.it} \\
   \And
  Pedro Neto \\
  CEMMPRE, Dep. Mechanical Engineering \\
  University of Coimbra \\
  Coimbra, Portugal\\
  \texttt{pedro.neto@dem.uc.pt} \\
}

\begin{document}
\maketitle

\begin{abstract}
Physical human-robot interaction (pHRI) remains a key challenge for achieving intuitive and safe interaction with robots. Current advancements often rely on external tactile sensors as interface, which increase the complexity of robotic systems. In this study, we leverage the intrinsic tactile sensing capabilities of collaborative robots to recognize digits drawn by humans on an uninstrumented touchpad mounted to the robot's flange. We propose a dataset of robot joint torque signals along with corresponding end-effector (EEF) forces and moments, captured from the robot's integrated torque sensors in each joint, as users draw handwritten digits (0–9) on the touchpad. The pHRI-DIGI-TACT dataset was collected from different users to capture natural variations in handwriting. To enhance classification robustness, we developed a data augmentation technique to account for reversed and rotated digits inputs. A Bidirectional Long Short-Term Memory (Bi-LSTM) network, leveraging the spatio-temporal nature of the data, performs online digit classification with an overall accuracy of 94\% across various test scenarios, including those involving users who did not participate in training the system. This methodology is implemented on a real robot in a fruit delivery task, demonstrating its potential to assist individuals in everyday life. Dataset and video demonstrations are available at: 
\url{https://TS-Robotics.github.io/pHRI-DIGI/}.
\end{abstract}


\section{Introduction}
Physical human-robot interaction (pHRI) has the potential to be a major enabler of natural interaction and collaboration between humans and robots in real-world scenarios \cite{DESANTIS2008253}. However, achieving intuitive and safe pHRI remains a significant challenge due to several factors, including safety concerns and the lack of reliable sensors targeting pHRI. Current collaborative robot safety directives establish operational guidelines, but in light of recent advancements, the naturalness and effectiveness of human-robot interaction still have significant room for improvement \cite{VILLANI2018248}. Most collaborative robots are capable of detecting contact with humans. However, physical interaction is typically limited to hand-guiding, and most applications focus primarily on preventing contact rather than enabling, classifying, and leveraging it.

\begin{figure}[t]
\centerline{\includegraphics[width=0.7\textwidth]{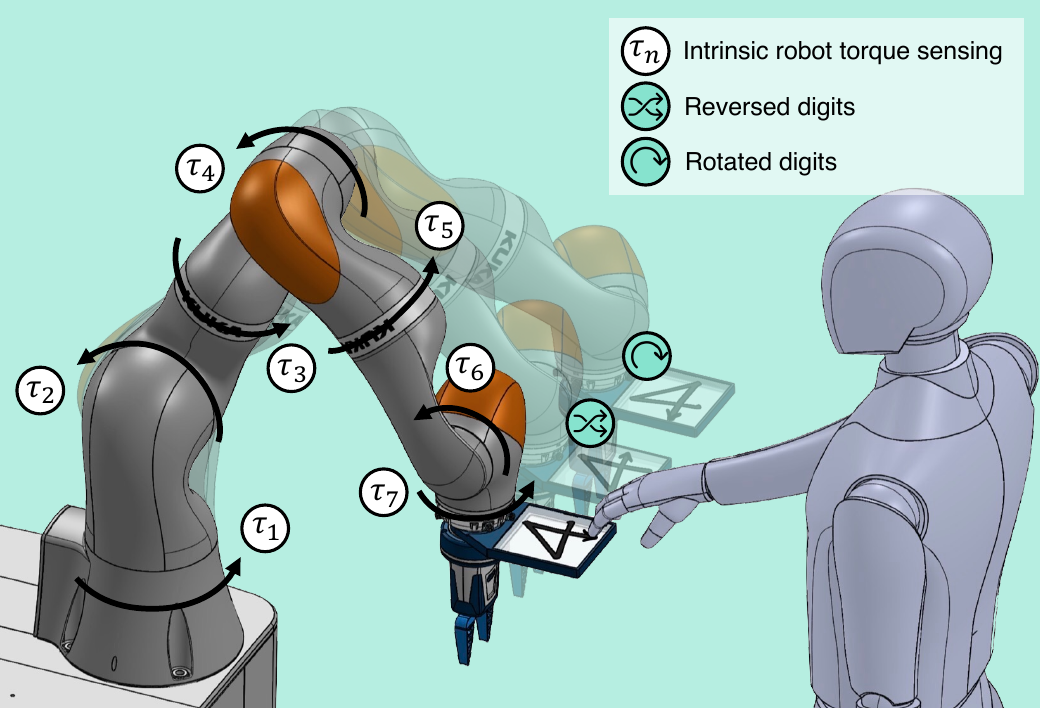}}
\caption{Conceptual overview of the proposed pHRI system: the system uses intrinsic robot torque sensing to recognize digits (0 - 9) drawn on a touchpad, including standard, reversed, and rotated variations.}
\label{FIG:systemoverview}
\end{figure}

\begin{figure*}[t]
\centerline{\includegraphics[width=\linewidth]{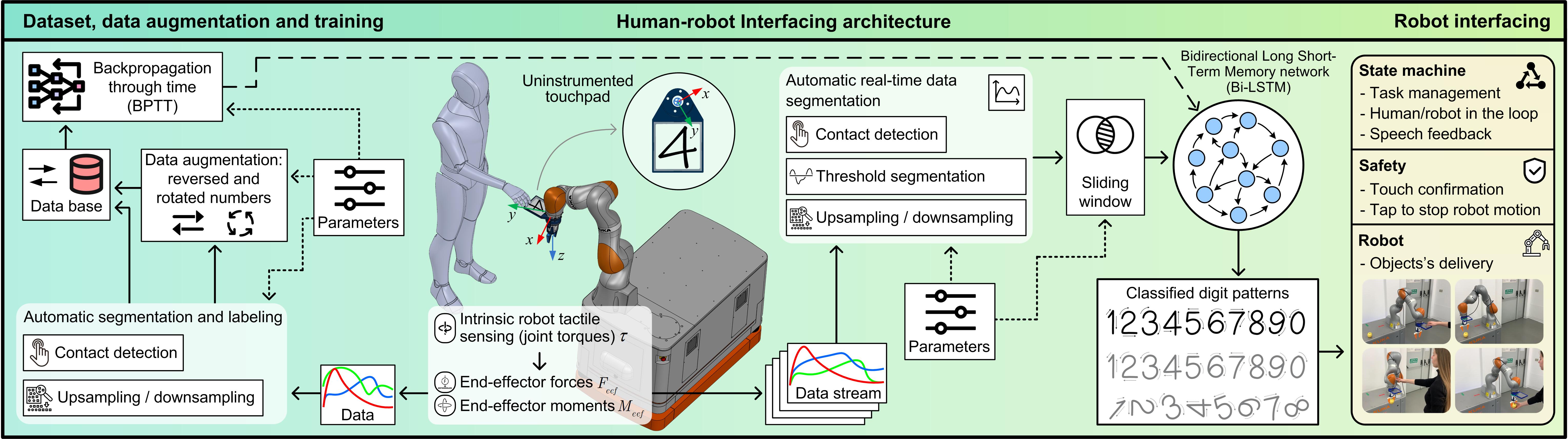}}
\caption{System architecture illustrating the data flow from the robot’s intrinsic joint torque sensing, through dataset collection and data augmentation, into the Bi-LSTM classification network. The classified patterns are then used to define the robot interface and guide the corresponding task execution.}
\label{FIG:systemarchitecture}
\end{figure*}

Over the past decades, force/torque sensors and touch/tactile sensing devices have been successfully used for pHRI tasks such as hand-guiding and touch detection \cite{4650764, Ajoudani}. More recently, these sensing technologies, combined with machine learning, have been effectively used as robotic electronic skin (e-skin) to detect and classify touch patterns. While these technologies have shown promise for enhancing pHRI, their integration complexity, along with limited mechanical flexibility and durability, poses challenges for seamless integration into robotic systems. 

In this context, considering both the limited applicability of the aforementioned sensing technologies and the integrated sensing capabilities of today’s collaborative robots, two important questions arise: 

\begin{enumerate}
    \item Are these sensing technologies truly necessary to enable meaningful pHRI? 
    \item Could we instead leverage the intrinsic tactile sensing already present in collaborative robots, such as joint torque measurements, to enable richer and more intuitive physical interactions beyond simple contact detection?
\end{enumerate}

Inspired by recent pioneering work on intrinsic robot sensing \cite{Iskandar2024}, this study leverages the intrinsic tactile sensing capabilities of collaborative robots to recognize digits drawn by humans, by sliding their finger, on an uninstrumented touchpad mounted to the robot’s flange, Fig. \ref{FIG:systemoverview}. The primary goal of this study is to assess whether the sensed joint torques, and the corresponding EEF forces and moments, are sufficiently sensitive to accurately classify the digits. We also investigate the generalizability of the classification model, specifically its online performance with users who did not participate in the training phase and with digits drawn in reversed or rotated orientations. A detailed overview of the system architecture, including the data processing pipeline and the Bi-LSTM network, is provided in Fig. \ref{FIG:systemarchitecture}. The key contributions of this paper are as follows:

\begin{itemize}
    \item An intuitive pHRI solution based on the intrinsic tactile sensing of collaborative robots: handwritten digits (0 - 9) are recognized by sliding the finger on an uninstrumented touchpad.
    \item Publicly available dataset pHRI-DIGI-TACT: it includes 10 classes of tactile handwritten digits (0–9) collected from multiple participants. Each class contains data on 7-DOF robot joint torques, EEF forces and moments.
    \item Enhanced robustness in classifying reversed and rotated digits: a data augmentation method enables the recognition of such variations when applied to the standard pHRI-DIGI-TACT dataset, increasing the system’s flexibility and general applicability.
    \item High-accuracy online digit classification: a bi-LSTM network achieves an overall accuracy of 94\% across various test scenarios, including those involving users who did not participate in the training phase.
    \item Real-world application: a robotic cell for object delivery, where the user interacts through the proposed interface to select the desired items. The robot is controlled via digits drawn on the touchpad, while the system provides feedback through text-to-speech voice output. These natural interaction methods demonstrate the intuitiveness of the interface, highlighting its potential to assist inexperienced users in everyday tasks.
\end{itemize}

\section{Related Work}
\label{SEC:relatedworks}
Physical human-robot interaction has gained increasing research interest in recent years due to its promising potential. Researchers are actively exploring both novel sensing technologies and methods for interpreting touch/tactile data within the robotics context, some of which powered by machine learning. The ability to sense and interpret physical contact between the robot’s structure and its surrounding environment, including interactions with humans, can open the door to new robotic applications and broaden accessibility for non-skilled users.

Promising approaches to enhancing pHRI involve equipping robots with external tactile sensors, such as e-skin, to provide high-resolution tactile feedback. Notable examples include a control framework utilizing high-resolution e-skin that covers the entire robot body \cite{Sun2024}, elastomeric e-skin with variable stiffness for safer human-robot interaction \cite{Park2022}, and robot e-skin solutions employing large-area tactile sensors to recognize human touch \cite{Aydin2024}. Soft and stretchable e-skin materials and their associated sensors offer adaptability to complex robot surfaces and are capable of integrating multimodal sensing, delivering high-resolution output, and providing extensive coverage \cite{doi:10.1126/scirobotics.aaz9239}. Although e-skin sensors offer multiple advantages, they are not yet commonly available in commercial robots due to several challenges. These include difficulties in sensor integration, ensuring long-term reliability, managing system complexity, achieving adaptability to varying environmental conditions, and providing the mechanical flexibility needed to conform to curved robot surfaces. Combining robotics with recent developments in e-skin technology and machine learning holds great promise for addressing these challenges and advancing the real-world applicability of pHRI.

A common approach involves mounting force/torque sensors on the robot’s flange to measure forces and torques at the robot EEF \cite{Jung2024, Aivaliotis02122018}. While practical, this method is limited in its ability to detect touch beyond the immediate vicinity of the EEF and can add bulk to the robot’s wrist. Alternative approaches have also been explored, such as using contact microphones to detect touch patterns \cite{GAMBOAMONTERO2023118510}.

Very few studies leverage the joint torque sensors already integrated into some collaborative robots for tactile and touch detection. These intrinsic sensors data, combined with robot dynamic models, can enable the estimation of contact points and the magnitude of forces along the robot structure \cite{Haddadin2017, Pang2021}.

Intrinsic robot joint torque sensing has been used in combination with machine learning to classify collision events into predefined categories \cite{Popov2017}. A recent pioneering work on intrinsic robot sensing further extends this approach by using machine learning to classify a rich set of robot contact interactions, including touch trajectories, letters, symbols, numbers, virtual buttons and slider bars \cite{Iskandar2024}. This system employs specialized robotic hardware featuring “sensing redundancy” (SARA, DLR, Germany), which integrates both motor- and link-side position sensors, as well as 6D force/torque sensors at the base and wrist of the robot.

Notably, no prior work has leveraged intrinsic robot joint torque sensing, available in some commercial collaborative robots, for digit classification aimed at enabling intuitive pHRI. As a result, our approach offers key advantages: (1) it utilizes intrinsic tactile sensing, eliminating the need for specialized robotic or sensing hardware, and (2) incorporates a robust digit classification model that achieves high accuracy, even with reversed or rotated digits.

\section{Methodology}
\label{SEC:methodology}
\subsection{System Overview and Setup}
In this study, we utilized a seven degrees of freedom collaborative robot (LBR iiwa 14 R820, KUKA, Germany) equipped with torque sensors at the link side of each joint, enabling precise measurement of joint torques during interaction. During dataset acquisition and fruit delivery task, we equipped the robot with a Schunk Co-Act collaborative gripper, enhanced with custom flexible fingers 3D-printed from Filaflex filament to ensure secure and compliant object grasping. The digit input interface consists of a custom-designed, uninstrumented touchpad, directly mounted onto the robot's flange, Fig \ref{FIG:systemoverview}. The touchpad is fabricated with a 3D-printed frame that securely holds a 120 mm x 120 mm x 10 mm plexiglass square. We selected plexiglass for its transparency, which prevents occlusions, visually reinforces the uninstrumented nature of the touchpad, and, for its smooth surface, facilitates easy finger movement during digit input.

Robot control, online classification of touch patterns, and data acquisition were managed from an external computer using MATLAB and the KUKA Sunrise Toolbox (KST) \cite{Safeea2019}. KST enabled the acquisition of joint torque signals, facilitated the online classification of the touch patterns, and enabled the transmission of motion commands.

\subsection{Data Collection}

\subsubsection{Digit Drawing Protocol}
For dataset consistency, the 0-9 digits were drawn following the patterns in the first row of Fig. \ref{FIG:DigitsProtocol}. The digit writing process on the touchpad is also shown in a video available in the supplementary material of this paper. Digits were drawn continuously, without lifting a finger, and users were instructed to complete each in approximately 2 seconds.
\begin{figure}[t]
\centerline{\includegraphics[width=0.7\textwidth]{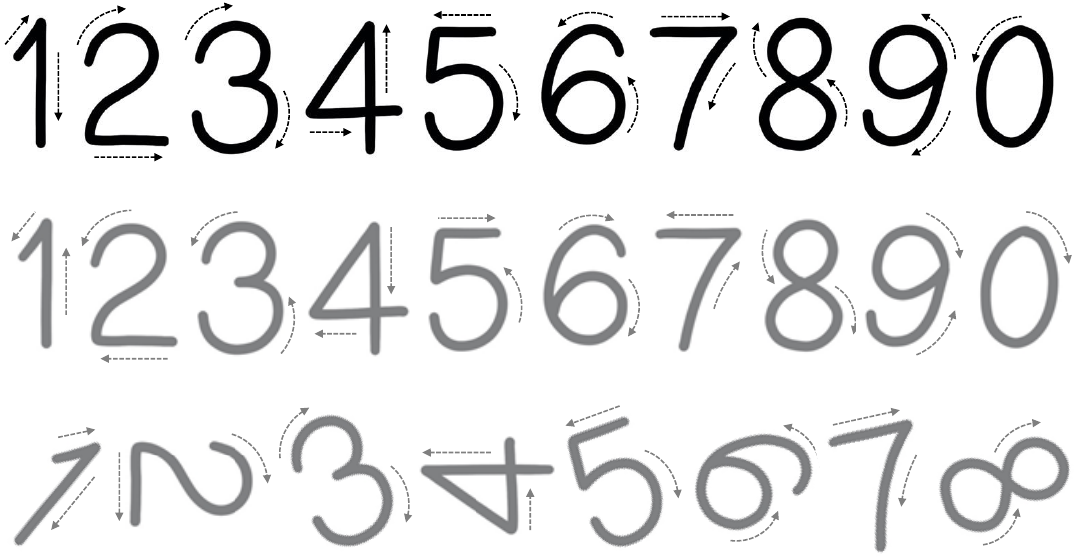}}
\caption{Illustration of digit drawing protocols: (Top) Standard digit drawing patterns used for data collection. (Middle) Reversed digit drawing patterns. (Bottom) Example of rotated digit drawing patterns.}
\label{FIG:DigitsProtocol}
\end{figure} 

\subsubsection{Data Acquisition and Offset Compensation}
During digit tracing, the robot's joint torque sensors measure the corresponding external joint torques $\bm{\tau}_{ext} \in \mathbb{R}^{7}$. In addition, the robot controller provides end-effector forces $\bm{F}_{ext}=(F_{x},F_{y},F_{z})$ and moments $\bm{M}_{ext}=(M_{x},M_{y},M_{z})$, calculated from the joint torques via the Jacobian. However, during preliminary tests, we observed that the raw joint torques and end-effector forces/moments exhibited offsets that varied between touches and across different robot poses. Because the torques generated during touchpad interaction are relatively small, failing to compensate for these offsets could negatively impact the classification accuracy. To mitigate these offsets, we implemented a sliding window technique for real-time baseline correction, using it both during dataset acquisition and for online classification.

\subsubsection{Dataset Generation, Segmentation, and Resampling}
To generate the dataset, we recorded continuous streams of digit tracing data. Isolating individual digits from these streams required a temporal segmentation process, achieved using a threshold-based approach. Touch initiation was detected when either the measured torque on joint 2 or the end-effector force along the z-axis exceeded a predetermined threshold. Similarly, termination of digit drawing was registered when the signal fell below the threshold. After isolating each touch pattern, we observed that the resulting number of acquired samples varied, due both to the non-real-time nature of Ethernet communication between the robot controller and MATLAB, and to slight variations in the time users took to draw the numbers. To ensure data consistency and facilitate effective training, the acquired signals for each segmented digit were resampled to a fixed length of 100 samples, employing either upsampling or downsampling techniques.

\subsubsection{Online Data Acquisition} 
A data acquisition process was implemented for online digit classification. The system continuously monitored the signals, employing a threshold-based approach to detect the initiation of a drawing digit (touch event). Upon detection, data was recorded for a predefined time window of fixed duration. Because this time window was chosen to accommodate more prolonged digit drawings, the recorded signal invariably contained a zero-padded segment at the end. To ensure consistency with the training dataset, this zero-padded segment was removed. In addition, to align with the training data and accommodate variations in drawing speed, the recorded signals were resampled to 100 samples using upsampling or downsampling techniques.

\subsection{Data Augmentation}
Even if the resulting geometric shape is the same, digits on the touchpad, can be drawn in a reversed manner: for example, the digit "2" is typically written from left to right and top to bottom following the standard protocol, but it could also be written from right to left and from bottom to top (see the second row of Fig. \ref{FIG:DigitsProtocol}). Similarly, digits could be drawn on the touchpad rotated by various angles, as shown in the third row of Fig. \ref{FIG:DigitsProtocol}. A video showing the numbers being drawn reversed and rotated is available in supplementary materials.

To classify reversed and rotated digits without collecting additional data, we implemented a data augmentation strategy using transformations of the existing dataset. While establishing direct relationships between joint torques is complex, modeling relationships between end-effector forces and moments is more straightforward.

To generate data for reversed digits, two factors must be considered: the reversal of the geometric path and the reversal of the finger direction. To account for the reversal of the geometric path, data corresponding to digits drawn in the standard way was flipped in time using a time reversal operator, denoted by $\mathcal{T}$. To account for the reversal of the finger direction, the signs of the force components along the x and y axes were inverted. Experimentally, we also observed that the signs of the moment components along the x and z axes should be inverted for reversed digits. This likely stems from the change in finger pressure distribution and orientation on the touchpad when drawing digits in reverse, which affects the measured moments. The moment along the y axis did not need to be inverted. Formally, given end-effector forces and moments for a digit drawn in the standard way, the augmented data representing the reversed digit, denoted with a tilde, is obtained as follows:
\begin{equation}
\begin{cases}
    \tilde{F}_{x}(t) = -\mathcal{T}(F_{x}(t)) \\
    \tilde{F}_{y}(t) = -\mathcal{T}(F_{y}(t)) \\
    \tilde{F}_{z}(t) = \mathcal{T}(F_{z}(t)) \\
\end{cases} \quad 
\begin{cases}
    \tilde{M}_{x}(t) = -\mathcal{T}(M_{x}(t)) \\
    \tilde{M}_{y}(t) = \mathcal{T}(M_{y}(t)) \\
    \tilde{M}_{z}(t) = -\mathcal{T}(M_{z}(t)) \\
\end{cases}.
\label{EQ:daya_augmentation_1}
\end{equation}
Similarly, to account for rotated digits by a certain angle $\theta$, the measured forces and moments were rotated by the same angle $\theta$ using a rotation matrix $R_z(\theta)$, representing rotation about the z-axis of the end-effector reference frame. Thus, the augmented data for rotated digits, denoted with a hat, is given by:
\begin{equation}
\begin{cases}
    \hat{F}_{x}(t) = R_z(\theta)F_{x}(t) \\
    \hat{F}_{y}(t) = R_z(\theta)F_{y}(t) \\
    \hat{F}_{z}(t) = R_z(\theta)F_{z}(t) \\
\end{cases} \quad 
\begin{cases}
    \tilde{M}_{x}(t) = R_z(\theta)M_{x}(t) \\
    \tilde{M}_{y}(t) = R_z(\theta)M_{y}(t) \\
    \tilde{M}_{z}(t) = R_z(\theta)M_{z}(t) \\
\end{cases}.
\label{EQ:data_augmentation_2}
\end{equation}

\subsection{Bi-LSTMs}
To classify the time-series data generated by digit tracing, we employed a Bidirectional Long Short-Term Memory (Bi-LSTM) network. LSTMs are a type of recurrent neural network (RNN) designed to effectively learn and classify sequential data by mitigating the vanishing gradient problem that can occur in traditional RNNs. This makes them well-suited for capturing the temporal dependencies inherent in the joint torque signals. The bidirectional architecture further enhances performance by processing the input sequence in both forward and reverse directions, allowing the network to capture contextual information from both past and future time steps relative to any given point in the sequence. This is particularly beneficial for digit recognition, as the shape of a digit is influenced by both preceding and subsequent movements.

\section{Experiments and results}
\label{SEC:experiments}

\subsection{Experimental Design}
The experiments were designed to evaluate the performance of the proposed digit recognition system under various conditions. The primary goals were to: 1) Evaluate the basic digit classification capability of the system, 2) Investigate the impact of training data source (single user vs. multiple users) on classification accuracy, 3) Assess the generalizability of the trained network to users who did not contribute to the training data, 4) Examine the robustness of the system to variations in robot pose, and 5) Evaluate the effectiveness of data augmentation for classifying reversed and rotated digits. These goals were pursued through two sets of experiments.

\begin{figure}[t]
\centerline{\includegraphics[width=0.9\textwidth]{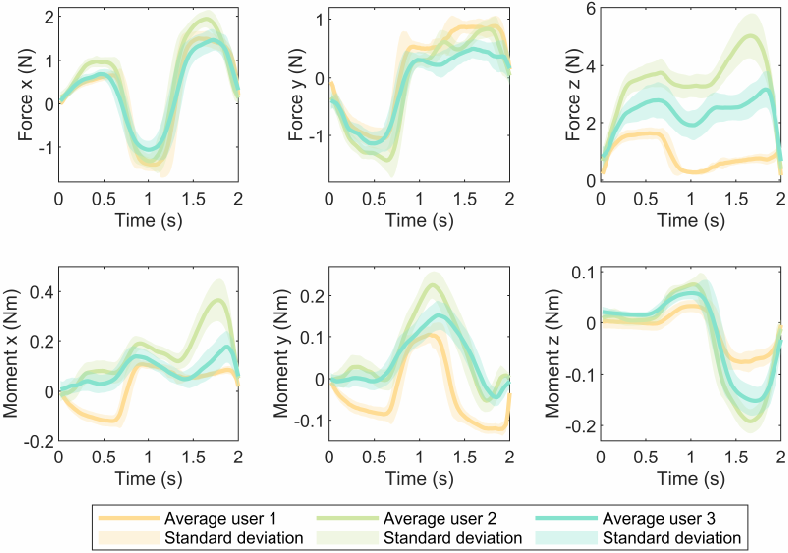}}
\caption{Mean end-effector forces and moments and
standard deviations during the drawing of the digit '4' by three different users. Data represents the average touch profile for 50 trials per user, revealing consistent trends.}
\label{FIG:SignalsN4}
\end{figure}

\subsubsection{Experiment Set 1: Baseline Performance and Generalizability}
The first set of experiments focused on evaluating the baseline performance and generalizability of the system using only normally drawn digits. A dataset was collected from a single user, containing 150 time series for each digit (0-9), for a total of 1500 time series. A second dataset was collected from three different users, with each user contributing 50 time series per digit, resulting in a dataset of 1500 time series. This ensured that the total size of the multi-user dataset matched that of the single-user dataset for fair comparison. During both data collection processes, the robot was maintained in a fixed home pose. The Cartesian coordinates of the home pose were (450, 0, 300) mm, and the Euler angles were (120$^\circ$, 0$^\circ$, -180$^\circ$). Only end-effector forces and moments were used as input features for training, while joint torques were not used. 

Separate networks were trained using the single-user and multi-user datasets. Prior to training, each dataset was split into training (70\%) and testing (30\%) subsets. All networks, across both Experiment Set 1 and Experiment Set 2, were trained using the Adam optimizer for 500 epochs, with a learning rate of 0.002 and 23 hidden layers. Once trained, the networks were tested using online classification. Online classification was first tested with the robot in its home pose, both by a user who contributed to the training data and by a user who did not. Each digit was drawn 10 times, and the classification accuracy was recorded. Online classification was then tested with the robot in 10 randomly selected poses within a defined range: $\Delta$x = $\pm$150 mm, $\Delta$y = $\pm$200 mm, $\Delta$z = $\pm$100 mm. The same testing procedure was used as in the home pose classification scenario.

\subsubsection{Experiment Set 2: Data Augmentation for Reversed and Rotated Digits}
The second set of experiments assessed the effectiveness of data augmentation techniques for classifying reversed and rotated digits, and to measure the impact of these data augmentation on the performance on non-augmented data. The multi-user dataset from Experiment Set 1 was augmented using the transformations described by Eq. \eqref{EQ:daya_augmentation_1} and \eqref{EQ:data_augmentation_2}. For rotated digits, rotation matrices along the z axis corresponding to angles of +90 degrees and -90 degrees were used. Separate networks were trained using the augmented datasets (one for reversed digits and one for rotated digits). Online classification was tested in a single robot pose by a single user who contributed to the training data. Each digit was drawn 10 times, and the classification accuracy was recorded.

\subsection{Results and Discussion}
To illustrate the magnitude of the recorded data, Fig. \ref{FIG:SignalsN4} shows the average end-effector forces and moments during the drawing of the digit '4' by different users, with shaded regions representing the standard deviation. The figure demonstrates that the forces and moments involved are relatively small. Specifically, the end-effector forces are typically on the order of 1-2 N, while the moments are generally less than 0.5 Nm. While the patterns of forces and moments are largely consistent across different users, the primary exception is the force in the z-direction, which varies depending on the contact pressure applied by each user. The low standard deviation, as shown in the graph, suggests a high degree of consistency in how users draw the digits.

\begin{figure}[t]
\centerline{\includegraphics[width=0.7\textwidth]{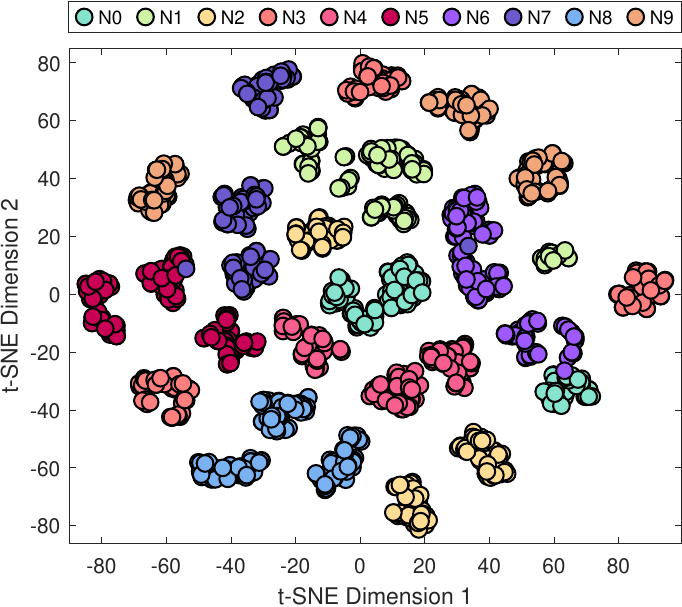}}
\caption{t-SNE plot showing the clustering of the multi-user dataset (1500 data points) in a reduced two-dimensional space, colored by digit class. The distinct clusters indicate high separability between digit classes.}
\label{FIG:tSNEgraph}
\end{figure}

\begin{figure}[t]
\centerline{\includegraphics[width=0.65\textwidth]{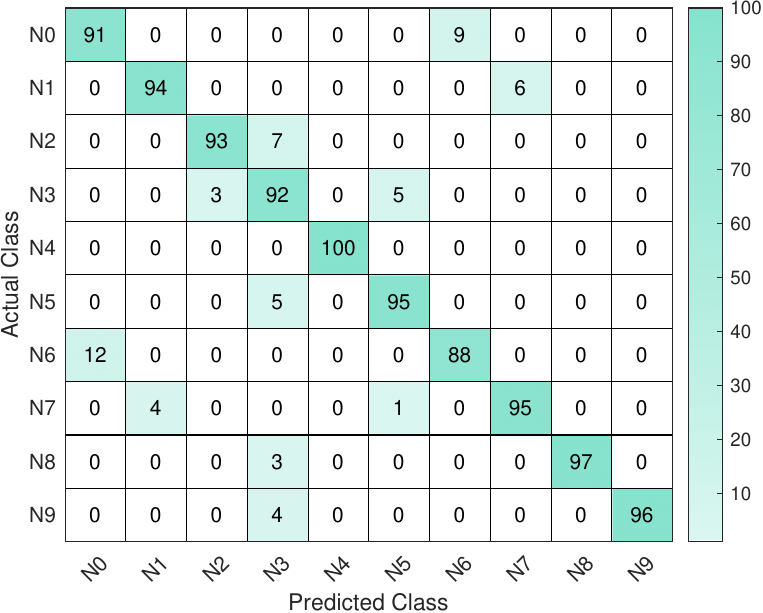}}
\caption{Confusion matrix showing the classification performance of the multi-user network when tested by a user who did not contribute to the training data, with the robot in variable poses.}
\label{FIG:ConfusionMatrix}
\end{figure}

The results from Experiment Set 1 are summarized in Table \ref{TAB:results1}. The offline accuracy was near 100\% for both datasets. Online classification with the robot in its home pose also yielded high accuracy when tested by a user who contributed to the training data. However, when tested by a user who did not contribute to the training data, the single-user network exhibited lower classification accuracy compared to the multi-user network. This highlights the need for multi-user datasets to achieve robust classification performance. To visualize the inherent separability of the digit classes within the multi-user dataset, we generated a t-SNE plot, Fig. \ref{FIG:tSNEgraph}. This plot reveals that data points corresponding to different digits form distinct clusters, indicating a clear separation between the classes. This separation supports the high classification accuracy observed in the experiments. Further analysis of the classification errors, as shown in the confusion matrix for online classification in different robot poses by a user who did not train the network, Fig. \ref{FIG:ConfusionMatrix}, reveals that the majority of errors occur between digits '0' and '6', '1' and '7', and '2' and '3'. The generalizability of the trained networks is supported by the consistency of the classification performance across different robot poses.

\begin{table}[htbp]
    \centering
    \caption{Classification accuracy for single-user and multi-user datasets.}
    \label{TAB:results1}
    \resizebox{\columnwidth}{!}{%
    \begin{tabular}{lccccccc}
    \toprule
         Dataset & Input features & Size & Offline & \multicolumn{2}{c}{Online (Home)} & \multicolumn{2}{c}{Online (Var. Poses)}   \\
         & & & Accuracy & Trained & Untrained & Trained & Untrained \\
         \midrule
         Single-user & $\bm{F}_{ext}$, $\bm{M}_{ext}$ & 1500 & 100\% & 99\% & 46\% & 97.4$\pm$1.64\% & 45.2$\pm$2.97\%\\
         Multi-user & $\bm{F}_{ext}$, $\bm{M}_{ext}$ & 1500 & 99.77\% & 100\% & 92\% & 96.9$\pm$1.19\% & 94.1$\pm$2.08\%\\
         \bottomrule
    \end{tabular}%
    }
\end{table}

The results from Experiment Set 2 are summarized in Table \ref{TAB:results2}. The offline accuracy remained near 100\%, and the classification of normally drawn digits was not negatively impacted by data augmentation. Reversed digits were classified with 92\% accuracy, with the primary source of error being the misclassification of digit '8' as digit '3'. Rotated digits were classified with an accuracy of approximately 80\%, with the most frequent errors occurring in the classification of digits '5', '1', and '7', indicating potential areas for improvement. It is also worth noting that the neural network trained only with normally drawn digits was able to classify numbers that were only slightly rotated (approximately less than 45 degrees).

\begin{table}[htbp]
    \centering
    \caption{Classification accuracy for augmented multi-user datasets.}
    \label{TAB:results2}
    \resizebox{\columnwidth}{!}{%
    \begin{tabular}{lccccccc}
        \toprule
        Dataset & Input features & Size & Offline & \multicolumn{4}{c}{Online Accuracy} \\
        & & & Accuracy & Normal & Reversed & +90$^{\circ}$ & -90$^{\circ}$ \\
        \midrule
         Multi-user, rev. aug. & $\bm{F}_{ext}$, $\bm{M}_{ext}$ & 3000 & 99.55\% & 99\% & 92\% & -- & -- \\
         Multi-user, rot. aug. & $\bm{F}_{ext}$, $\bm{M}_{ext}$ & 4500 & 98.77\% & 100\% & -- & 86\% & 81\% \\
         \bottomrule
    \end{tabular}%
    }
\end{table}

\subsection{Real-world application}
\begin{figure}[t]
\centerline{\includegraphics[width=0.75\textwidth]{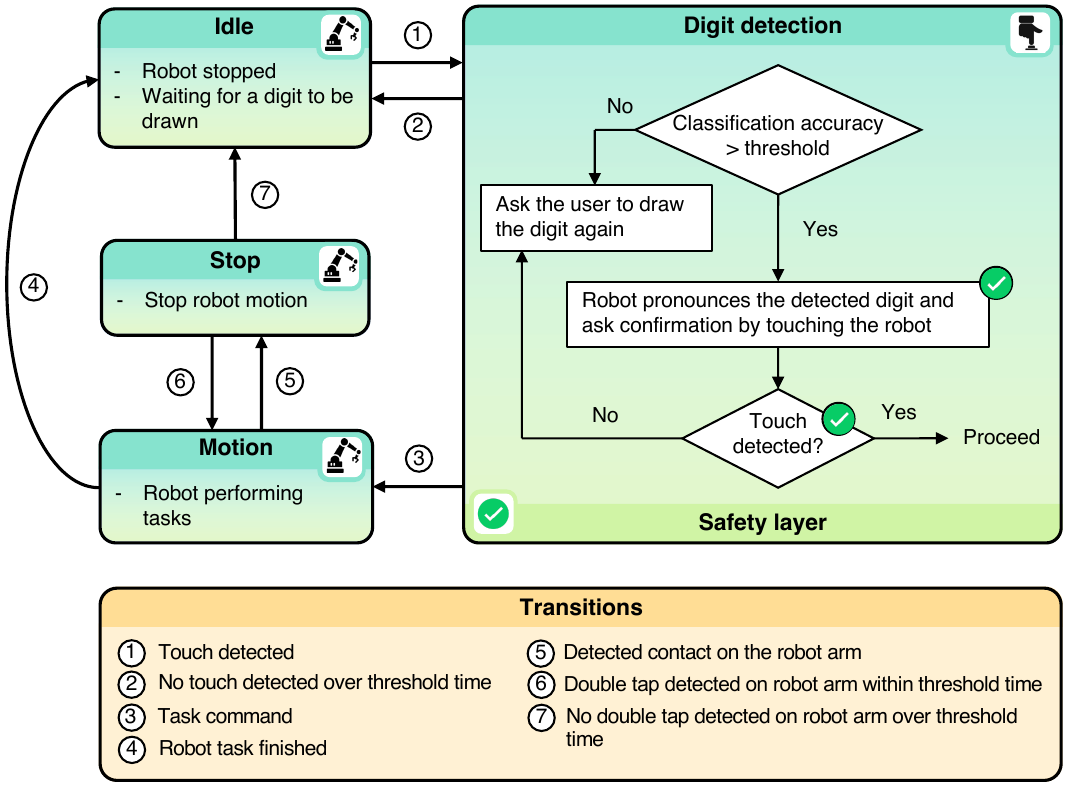}}
\caption{Hierarchical Finite State Machine (HFSM) for digit-based task execution. The HFSM governs the robot's behavior, managing task execution, safety protocols, and user interaction. The system utilizes spoken feedback and touch confirmation to enable effective digit recognition and task execution.}
\label{FIG:statemachine}
\end{figure}

To facilitate the practical deployment of the touchpad in real-world applications, we developed a Hierarchical Finite State Machine (HFSM) for task management and safety, Fig. \ref{FIG:statemachine}.

Within this HFSM framework, each digit drawn on the touchpad is assigned to a specific task. The system initiates in the \textit{Idle} state, awaiting user input. Upon detection of a touch event (transition 1), the system transitions to the \textit{Detection of digit} state. A safety layer is implemented within this state to ensure reliable digit classification. If the neural network's classification confidence score falls below a pre-defined threshold, the system verbally prompts the user to redraw the digit. Notably, the system provides intuitive feedback and instructions to the user via synthesized speech, a more accessible interface than traditional Graphical User Interfaces (GUIs). If the classification confidence exceeds the threshold, the system audibly confirms the recognized command to the user and requests confirmation by touch on the robot. Upon successful touch detection, the system proceeds to execute the corresponding action and transition to the \textit{Motion} state (transition 3), otherwise, the system returns to \textit{Idle} state. For enhanced safety during robot motion, touch detection on the robot arm while in the \textit{Motion} state triggers a transition to the \textit{Stop} state (transition 5). A subsequent double-tap touch event detected within a predefined time window returns the system to the \textit{Motion} state (transition 6), otherwise, the system transitions back to the \textit{Idle} state (transition 4).

\begin{figure*}[ht!]
\centerline{\includegraphics[width=\linewidth]{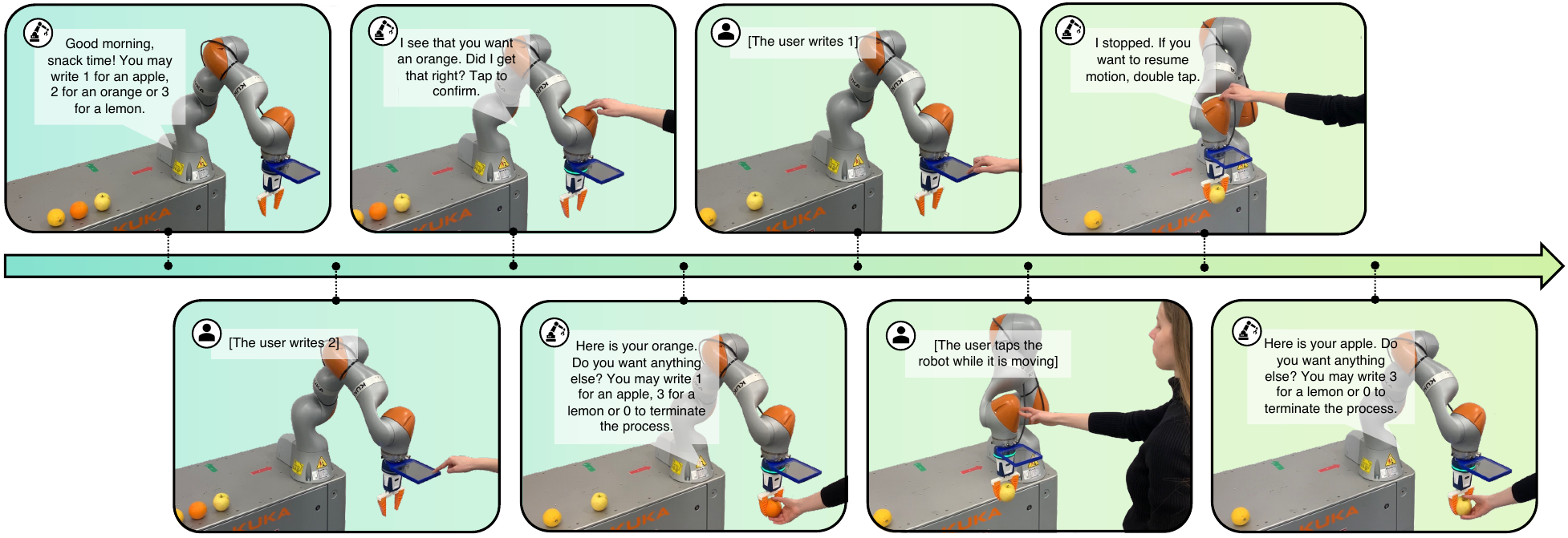}}
\caption{Snapshots of the Fruit Delivery Task. The robot verbally guides the user and delivers the fruit depending on the digit that is drawn (in this case 1 is for apple, 2 is for orange and 3 is for lemon).}
\label{FIG:demo}
\end{figure*}

To validate the practical applicability of the drawing digits system, we implemented a fruit delivery use case. In this scenario, different digits are assigned to different fruit items, enabling users to request specific items via the touchpad interface. The robot then autonomously retrieves and presents the requested fruit to the user. A video demonstration of the system is available in the supplementary materials accompanying this work. A set of snapshots of the demo are reported in Fig. \ref{FIG:demo}.

\section{Conclusion}
\label{SEC:conclusion}
This study has demonstrated the feasibility of leveraging intrinsic joint torque sensor data from collaborative robots to effectively classify digits drawn on an instrumented touchpad by a human user. The results show that the proposed system can achieve high accuracy in recognizing digits, even when the input data is augmented to account for reversed and rotated numbers. This approach offers a promising alternative to traditional pHRI methods that rely on external tactile sensors, reducing system complexity and cost. The real-world demonstration of a fruit delivery task showcases the potential for intuitive physical interaction between humans and robots using the proposed digit-based interface. Future work will focus on expanding the robustness and versatility of the system. This includes exploring the classification of digits drawn following different geometric paths, with pauses, and extending the system to recognize multi-digit numbers.

\section*{Acknowledgments}
This research is sponsored by national funds through FCT Fundação para a Ciência e a Tecnologia, under the projectUIDB/00285/2020 and LA/P/0112/2020.

\bibliographystyle{unsrt}  
\bibliography{references}

\begin{thebibliography}{10}

\bibitem{DESANTIS2008253}
Agostino {De Santis}, Bruno Siciliano, Alessandro {De Luca}, and Antonio
  Bicchi.
\newblock An atlas of physical human–robot interaction.
\newblock {\em Mechanism and Machine Theory}, 43(3):253--270, 2008.

\bibitem{VILLANI2018248}
Valeria Villani, Fabio Pini, Francesco Leali, and Cristian Secchi.
\newblock Survey on human–robot collaboration in industrial settings: Safety,
  intuitive interfaces and applications.
\newblock {\em Mechatronics}, 55:248--266, 2018.

\bibitem{4650764}
Sami Haddadin, Alin Albu-Schaffer, Alessandro De~Luca, and Gerd Hirzinger.
\newblock Collision detection and reaction: A contribution to safe physical
  human-robot interaction.
\newblock In {\em 2008 IEEE/RSJ International Conference on Intelligent Robots
  and Systems}, pages 3356--3363, 2008.

\bibitem{Ajoudani}
Arash Ajoudani, Andrea~Maria Zanchettin, Serena Ivaldi, Alin Albu-Schaffer,
  Kazuhiro Kosuge, and Oussama Khatib.
\newblock Progress and prospects of the human–robot collaboration -
  autonomous robots, Oct 2017.

\bibitem{Iskandar2024}
Maged Iskandar, Alin Albu-Schäffer, and Alexander Dietrich.
\newblock Intrinsic sense of touch for intuitive physical human-robot
  interaction.
\newblock {\em Science Robotics}, 9(93), 2024.

\bibitem{Sun2024}
Yu~Sun, Cong Xiao, Lipeng Chen, Lu~Chen, Haojian Lu, Yue Wang, Wang-Wei Lee,
  Yu~Zheng, Zhengyou Zhang, and Rong Xiong.
\newblock Beyond end-effector: Utilizing high-resolution tactile signals for
  physical human-robot interaction.
\newblock {\em IEEE Transactions on Industrial Electronics}, 2024.

\bibitem{Park2022}
K.~Park, H.~Yuk, M.~Yang, J.~Cho, H.~Lee, and J.~Kim.
\newblock A biomimetic elastomeric robot skin using electrical impedance and
  acoustic tomography for tactile sensing.
\newblock {\em Science Robotics}, 7(67), 2022.

\bibitem{Aydin2024}
Ahmet Aydin and Erdinç Avaroğlu.
\newblock Contact classification for human–robot interaction with densely
  connected convolutional neural network and convolutional block attention
  module.
\newblock {\em Signal, Image and Video Processing}, 18(5):4363 – 4374, 2024.

\bibitem{doi:10.1126/scirobotics.aaz9239}
Benjamin Shih, Dylan Shah, Jinxing Li, Thomas~G. Thuruthel, Yong-Lae Park,
  Fumiya Iida, Zhenan Bao, Rebecca Kramer-Bottiglio, and Michael~T. Tolley.
\newblock Electronic skins and machine learning for intelligent soft robots.
\newblock {\em Science Robotics}, 5(41):eaaz9239, 2020.

\bibitem{Jung2024}
Dawoon Jung, Seongun Bu, and Uikyum Kim.
\newblock Arbitrary surface contact sensing method for physical human-robot
  interaction.
\newblock {\em IEEE Transactions on Industrial Informatics}, 20(6):8274 –
  8283, 2024.

\bibitem{Aivaliotis02122018}
P.~Aivaliotis, G.~Michalos, and S.~Makris and.
\newblock Cooperating robots for fixtureless assembly: modelling and simulation
  of tool exchange process.
\newblock {\em International Journal of Computer Integrated Manufacturing},
  31(12):1235--1246, 2018.

\bibitem{GAMBOAMONTERO2023118510}
Juan~Jose Gamboa-Montero, Fernando Alonso-Martin, Sara Marques-Villarroya, Joao
  Sequeira, and Miguel~A. Salichs.
\newblock Asynchronous federated learning system for human–robot touch
  interaction.
\newblock {\em Expert Systems with Applications}, 211:118510, 2023.

\bibitem{Haddadin2017}
Sami Haddadin, Alessandro De~Luca, and Alin Albu-Schäffer.
\newblock Robot collisions: A survey on detection, isolation, and
  identification.
\newblock {\em IEEE Transactions on Robotics}, 33(6):1292 – 1312, 2017.

\bibitem{Pang2021}
Tao Pang, Jack Umenberger, and Russ Tedrake.
\newblock Identifying external contacts from joint torque measurements on
  serial robotic arms and its limitations.
\newblock {\em Proceedings - IEEE International Conference on Robotics and
  Automation}, 2021-May:7665 – 7671, 2021.

\bibitem{Popov2017}
Dmitry Popov, Alexandr Klimchik, and Nikolaos Mavridis.
\newblock Collision detection, localization and classification for industrial
  robots with joint torque sensors.
\newblock {\em RO-MAN 2017 - 26th IEEE International Symposium on Robot and
  Human Interactive Communication}, 2017-January:838 – 843, 2017.

\bibitem{Safeea2019}
M.~{Safeea} and P.~{Neto}.
\newblock Kuka sunrise toolbox: Interfacing collaborative robots with matlab.
\newblock {\em IEEE Robotics Automation Magazine}, 26(1):91--96, March 2019.

\end{thebibliography}

\end{document}